\documentclass[conference]{IEEEtran}

\usepackage{cite}
\usepackage{amsmath,amssymb}
\usepackage{graphicx}
\usepackage{float}
\usepackage[table]{xcolor}
\usepackage{url}
\usepackage{array}
\usepackage{booktabs}
\usepackage{multirow}
\usepackage{hyperref}
\usepackage{eso-pic}

\usepackage[font=footnotesize,skip=2pt]{caption}

\IEEEoverridecommandlockouts

\makeatletter
\let\COMPASorigtitlepagestyle\ps@IEEEtitlepagestyle
\def\ps@IEEEtitlepagestyle{%
    \COMPASorigtitlepagestyle
    \def\@oddhead{%
        \hfill
        \parbox{0.62\textwidth}{%
            \raggedleft\scriptsize
            IEEE 3rd International Conference on Computing, Applications and Systems (COMPAS 2026)\\
            9--10 October 2026, University of Dhaka, Bangladesh
        }%
    }%
    \let\@evenhead\@oddhead
}
\makeatother

\begin{document}
\AddToShipoutPictureFG*{%
    \AtPageLowerLeft{%
        \put(\LenToUnit{0.63in},\LenToUnit{0.45in}){%
            \scriptsize
            979-8-3195-3773-7/26/\$31.00~\copyright~2026 IEEE
        }%
    }%
}

\title{LUMO (Lightweight Unified Multilingual Orchestrator): A Privacy Preserving Offline Voice Assistant}

\author{
\IEEEauthorblockN{
Md. Mehedi Hasan Naeem,
Mst. Kamrunnahar Ruma,
Nafiza Anjum,
Shakila Sultana,
Md. Sujan Ali
}
\IEEEauthorblockA{
\textit{Department of Computer Science and Engineering} \\
\textit{Jatiya Kabi Kazi Nazrul Islam University} \\
Mymensingh, Bangladesh \\
\texttt{mehedinaeem00@gmail.com},
\texttt{kamrunnaharruma242@gmail.com} \\
\texttt{nafizaanjum.2002@gmail.com},
\texttt{shakila161470@gmail.com},
\texttt{sujan\_cse@jkkniu.edu.bd}
}
}


\maketitle

\begin{abstract}
Reliable voice interaction is essential in environments with limited internet connectivity and strong privacy. However, most existing voice assistants depend on cloud-based services, which leads to latency issues, dependency on internet access, and privacy vulnerabilities. This research presents LUMO (Lightweight Unified Multilingual Orchestrator), a privacy preserving offline voice assistant designed for edge computing environments. This system integrates local Automatic Speech Recognition (ASR), locally deployed quantized Large Language Model (LLM), and Text-to-Speech (TTS) synthesis into a fully offline pipeline running on a Raspberry Pi 5 with 8\,GB RAM.
To enable efficient operation on resource constrained hardware, the language model is compressed using 4-bit GGUF quantization, which reduces memory usage while preserving practical conversational capability. Existing edge based voice assistants Mycroft provides partial offline functionality without a generative LLM, with an approximate latency of $\sim$5\,s and power consumption of $\sim$12\,W, while Rhasspy supports full offline operation but lacks generative capabilities, with $\sim$3\,s latency and $\sim$11\,W power usage. In contrast, LUMO achieves a Word Error Rate (WER) of 6.8\% for short English utterances in low noise conditions, an end-to-end response latency of 2.0--4.0\,s, and a lower peak power consumption of approximately 9.0\,W.The system also achieves effective offline recognition for Bangla speech, supporting multilingual accessibility in low resource settings. By operating entirely offline, LUMO provides strong data privacy, reduced need for cloud connectivity, and suitability for privacy sensitive edge execution such as rural healthcare, education, and disaster response scenarios.

\end{abstract}


\section{Introduction}

Voice assistant have become an important part of human computer interaction, enabling applications in smart homes, healthcare, education and assistive technology. However, most commercial systems use cloud infrastructure for speech recognition, language understanding, and response generation.This dependency introduces network requirements, latency and privacy concerns due to the transmission of user data to remote servers. Recent advances in edge AI and lightweight language models enable intelligent applications on resource constrained devices.TinyLlama provides language reasoning with a relatively small number of parameters [1]. Local ASR enables speech recognition without internet connection [2]. Model quantization reduces the memory and computational requirements and has practical performance [3].These developments are making fully offline voice assistants more real than ever. Despite this progress, it is still challenging to deploy conversational assistants on edge hardware.Lightweight models need to balance inference speed, memory consumption and response quality. Rule-based intent recognition is used by many existing offline assistants, which limits theirability to support open-ended conversation.Generative models like LLaMA allow for more flexible, context-aware interaction [4], but many studies focus on individual models rather than full offline systems that integrate speech recognition, language reasoning, and speech synthesis.Also, system-level evaluation is important for practical edge deployment.Metrics like response latency, power consumption, memory usage, and thermal stability are directly impactful in real-world operation, yet are often less emphasized than model-level benchmarks [5].To tackle these challenges, this paper proposes LUMO (Lightweight Unified Multilingual Orchestrator), a privacy-preserving offline voice assistant for Raspberry Pi devices.LUMO is a combination of local automatic speech recognition, 4-bit GGUF-quantized lightweight language model, and offline text-to-speech synthesis in an 8GB memory environment.The system supports interaction in English and Bangla and is evaluated on the basis of speech recognition accuracy, response latency, memory usage, power consumption, and thermal performance. Section II presents related work. Section III presents system design and methodology. Section IV describes experimental setup. Section V discusses results. Section VI concludes the paper.

\section{Background Study and Related Work}

The cloud-based voice assistants offer strong computational power but are hampered by latency, internet dependency and privacy issues. Edge computing addresses these issues by processing data near the source, reducing communication delay and bandwidth requirements \cite{shi2016edge,satyanarayanan2017edge}. Recent studies have further explored efficient AI deployment on resource-constrained edge platforms \cite{chen2023edgeai,li2024edgeai}.

Raspberry Pi and similar embedded devices have been widely studied for lightweight AI applications \cite{uthayakumar2022raspberry}. Techniques for efficient deep-learning execution on mobile and embedded hardware have also been proposed \cite{lane2017squeezing,verhelst2023edge}. However, deploying large language models on such devices remains difficult because of their high memory and computational requirements. While LLaMA and LLaMA 2 demonstrated strong transformer-based language capabilities \cite{touvron2023llama,touvron2023llama2}, lightweight models such as TinyLlama offer more practical alternatives for edge deployment \cite{zhang2024tinyllama,gupta2024tinytransformers}. Quantization methods including GPTQ and QLoRA further reduce memory and computation requirements \cite{frantar2023gptq,dettmers2023qlora}, while LLMPi demonstrates the feasibility of optimized LLM inference on Raspberry Pi devices \cite{llmpi2025}.

Speech processing is another essential component of offline voice assistants. Kaldi provides a widely used ASR framework \cite{povey2011kaldi}, while voice activity detection improves real-time speech processing efficiency \cite{tan2020webrtc}. Research has also addressed offline ASR and robustness in noisy environments \cite{ko2023offlineasr,li2023robustspeech,kim2022robustspeech}. Multilingual resources such as Common Voice \cite{ardila2020commonvoice}, multilingual ASR models \cite{pratap2020massively}, and Whisper \cite{radford2023whisper} have further improved multilingual speech recognition. VOSK provides efficient offline recognition for embedded platforms \cite{alphacephei2024vosk}.

Although embedded voice assistants have been proposed \cite{lazzaroni2024embedded}, many systems still rely on rule-based command processing and provide limited open-ended conversational capability. Existing lightweight-LLM studies also mainly emphasize model efficiency rather than complete offline voice-assistant integration, while practical factors such as latency, thermal behavior, and power consumption are often insufficiently evaluated.

To address these gaps, this work introduces LUMO (Lightweight Unified Multilingual Orchestrator), a privacy-preserving offline voice assistant for Raspberry Pi. LUMO integrates a quantized lightweight language model, on-device speech recognition, and text-to-speech synthesis into a unified multilingual architecture designed for efficient operation on resource-constrained hardware.

\section{Proposed System Design and Methodology}

This section presents the architecture and implementation of LUMO, which integrates speech processing, language reasoning, and speech synthesis within a fully offline edge framework.

\subsection{System Overview}

LUMO is a modular pipeline of Voice Activity Detection (VAD), Speech-to-Text (STT), local Large Language Model (LLM), and Text-to-Speech (TTS) based on Raspberry Pi 5. It detects and transcribes the speech. The local LLM processes it and converts it to audio to play back.

\begin{figure}[h]
\centering
\includegraphics[width=0.80\columnwidth]{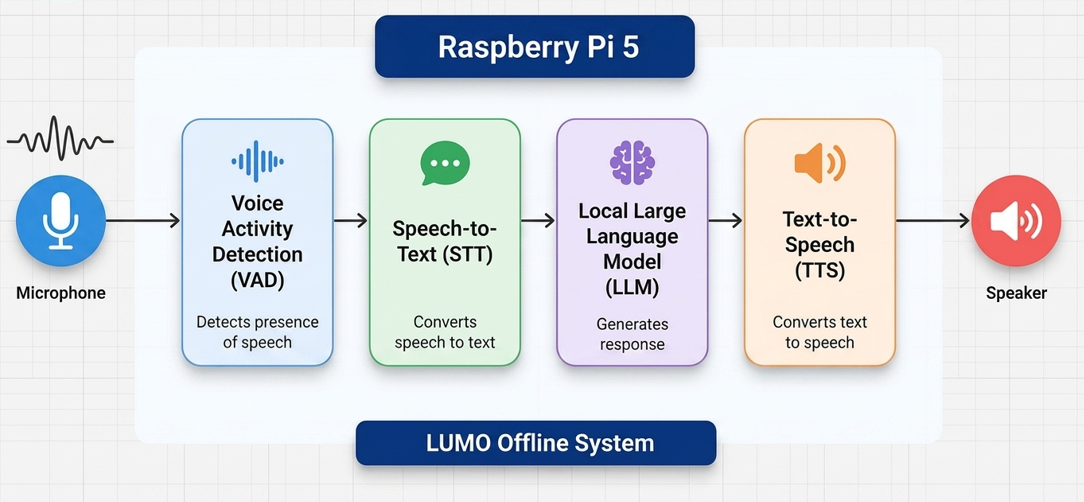}
\caption{Overall architecture of the proposed LUMO offline voice assistant system running on Raspberry Pi 5.}
\label{fig:lumo_architecture}
\end{figure}

All the processing is done locally as shown in Fig.~\ref{fig:lumo_architecture}, which helps preserve user privacy, reduces dependence on network, and enables low-latency interaction \cite{shi2016edge}.

\subsection{Multilingual Speech Dataset Design}

A custom multilingual dataset was developed to evaluate speech recognition using short command-style English and Bangla utterances, such as ``play meditation music'' and ``dhyaner gaan chalu koro.'' Ten bilingual speakers (3 male and 7 female) recorded manually transcribed utterances using the same USB condenser microphone and deployment environment. Since pretrained ASR and language models were used without task-specific fine-tuning, the dataset served only for evaluation.

\subsection{Hardware Configuration}

LUMO runs on a Raspberry Pi 5 with 8~GB RAM, selected for its low cost, flexibility, and Linux compatibility \cite{uthayakumar2022raspberry}. A USB condenser microphone and speaker provide audio input and output, while all processing is performed locally. Raspberry Pi supports optimized edge-AI inference frameworks \cite{verhelst2023edge}. A NeoPixel LED ring indicates listening, processing, and speaking states.

\subsection{Software Pipeline}

LUMO is implemented in Python on a Linux-based operating system and performs the complete processing pipeline locally.

\begin{figure}[h]
\centering
\includegraphics[width=0.9\columnwidth]{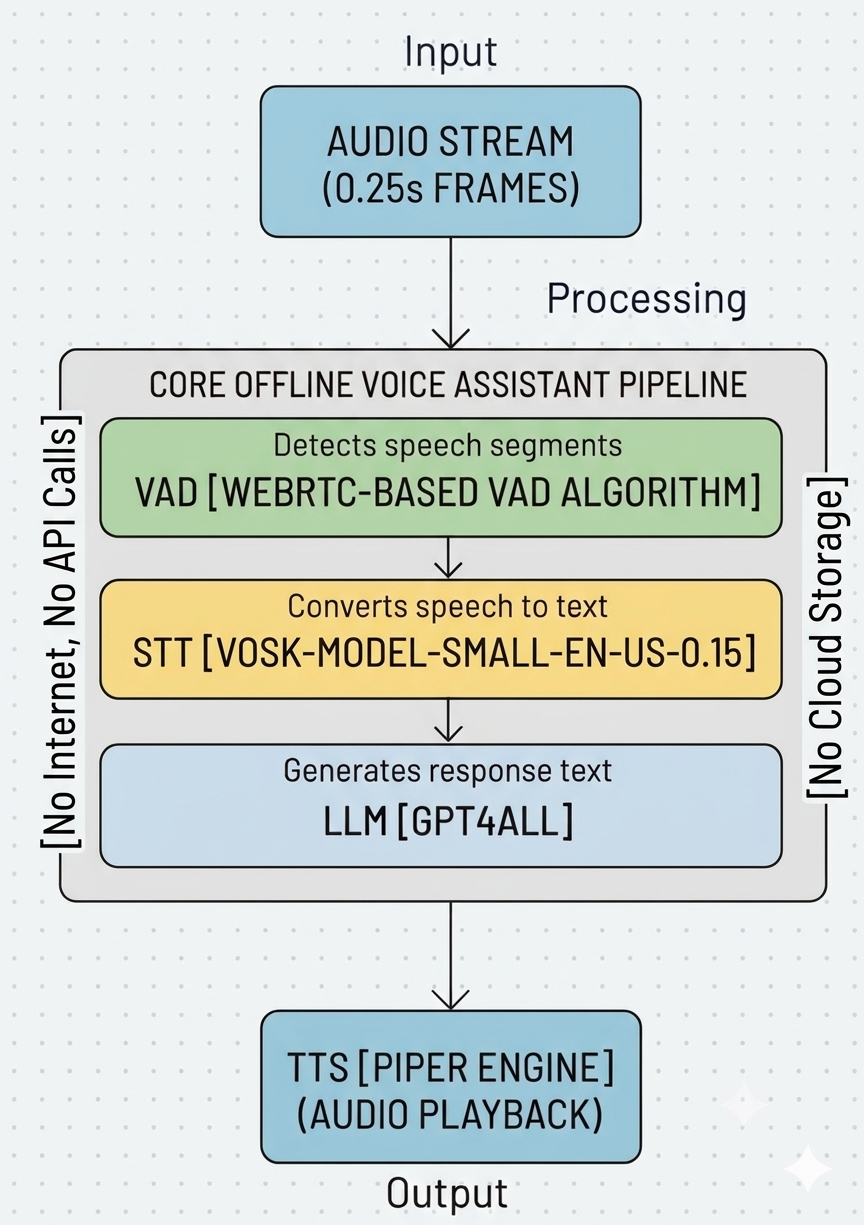}
\caption{Software processing pipeline for the LUMO voice assistant system.}
\label{fig:software_pipeline}
\end{figure}

As shown in Fig.~\ref{fig:software_pipeline}, audio is captured in approximately 0.25-second frames and analyzed by VAD. Detected speech is transcribed, processed by the LLM, and converted into speech by the TTS module. The modular design enables independent component optimization and efficient edge deployment \cite{lane2017squeezing}.

\subsection{Voice Activity Detection and Speech Recognition}

WebRTC-based VAD identifies speech frames before transcription, reducing unnecessary processing \cite{tan2020webrtc}. Detected speech is transcribed using the VOSK offline ASR framework based on Kaldi, which provides efficient speech recognition for resource-constrained devices \cite{povey2011kaldi,ko2023offlineasr}.

\subsection{Local LLM Integration}

Recognized text is passed to a locally deployed LLM for response generation. LUMO uses GGUF-quantized models with lower-precision weights to reduce memory requirements while preserving language understanding \cite{dettmers2023qlora}. Local inference provides context-aware responses without transmitting user data to cloud services.

\subsection{Text-to-Speech Module}

LUMO uses Piper TTS to convert generated responses into speech with low computational requirements.

\begin{figure}[h]
\centering
\includegraphics[width=0.8\columnwidth]{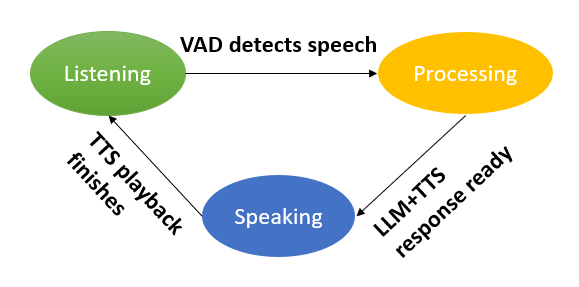}
\caption{Operational state transitions of LUMO during voice interaction.}
\label{fig:state_diagram}
\end{figure}

Fig.~\ref{fig:state_diagram} shows the listening, processing, and speaking states. Offline TTS engines such as Piper and Coqui TTS support natural speech generation without internet connectivity \cite{casanova2024coqui}.

\subsection{Concurrency and Control Logic}

\begin{figure}[h]
\centering
\includegraphics[width=0.8\columnwidth]{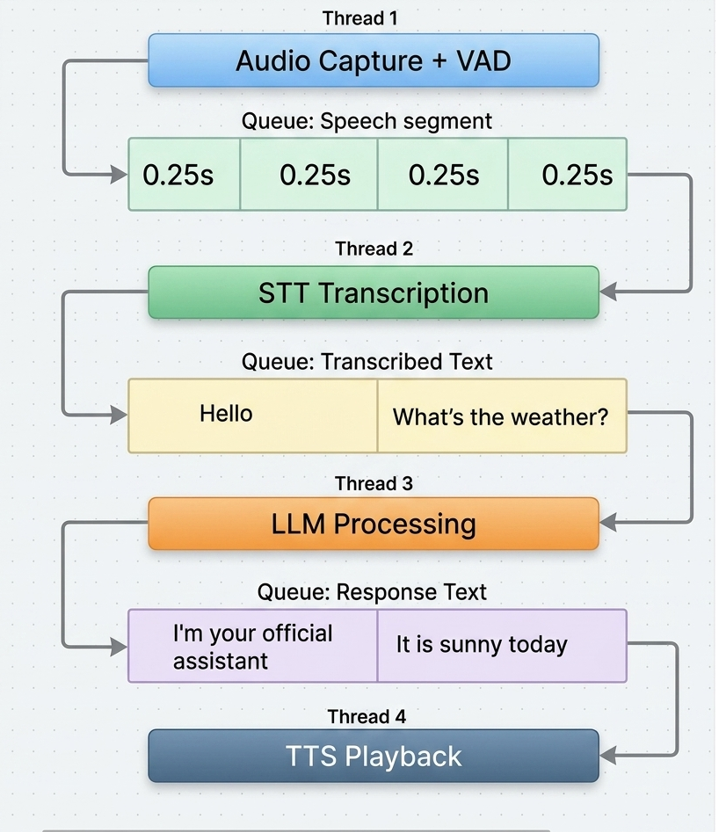}
\caption{Multithreaded processing architecture used in the LUMO system.}
\label{fig:thread_architecture}
\end{figure}

LUMO uses separate threads for audio capture, speech recognition, LLM inference, and playback. As shown in Fig.~\ref{fig:thread_architecture}, audio and text are passed between modules through queues, reducing blocking and supporting low-latency real-time interaction \cite{li2022speechsystems}.

\subsection{Deployment Optimization}

Quantization reduces memory and computational requirements, while short audio frames reduce processing latency. USB SSD storage improves model loading speed, and active cooling maintains stable temperature during prolonged inference \cite{chen2023efficientedgeai}.
\section{Experimental Setup}

LUMO was evaluated in a fully offline edge environment using speech recognition accuracy, response latency, memory usage, power consumption, and thermal behavior as the primary performance metrics. The experiments were designed to reflect realistic operation on resource-constrained hardware without internet connectivity.

\subsection{Hardware and Software Environment}

LUMO was evaluated on a Raspberry Pi 5 with an ARM Cortex-A76 processor, 8~GB RAM, and 64-bit Raspberry Pi OS. A USB condenser microphone and standard speaker were used for audio input and output. The Python 3.10 implementation integrates VOSK for offline speech recognition, WebRTC for voice activity detection, GPT4All for quantized local LLM inference, and Piper for offline text-to-speech synthesis.

\subsection{Speech Dataset}

A custom bilingual speech dataset containing 1,000 command-style utterances was developed to evaluate LUMO's offline speech recognition. Ten bilingual speakers recorded 500 English and 500 Bangla utterances, each containing 3--10 words and representing typical voice-assistant interactions.

\begin{table}[h]
\centering
\caption{LUMO Speech Dataset Summary}
\label{tab:dataset_summary}
\footnotesize
\setlength{\tabcolsep}{4pt}
\renewcommand{\arraystretch}{1.0}
\begin{tabular}{p{0.31\columnwidth} p{0.61\columnwidth}}
\hline
\textbf{Attribute} & \textbf{Value} \\
\hline
Languages & English, Bangla \\
Speakers & 10 bilingual speakers \\
Utterances & 1,000 (500 per language) \\
Command Length & 3--10 words \\
Audio & 16-kHz mono WAV \\
Recording Device & USB condenser microphone \\
Annotation & Manually verified transcripts \\
Metadata & Speaker ID, language, transcript (CSV) \\
\hline
\end{tabular}
\end{table}

Recordings were captured using the same USB condenser microphone used in the deployment setup and stored as 16-kHz mono WAV files. Ground-truth transcriptions were manually verified, while speaker ID, language, and transcript information were stored as CSV metadata. The dataset and LUMO implementation are publicly available at \url{https://github.com/mehedinaeem/lumo-speech-dataset} and \url{https://github.com/mehedinaeem/Lumo}, respectively.

\subsection{Noise-Level Test Conditions}

Speech recognition robustness was evaluated under three controlled ambient noise conditions: quiet ($<30$\,dB), moderate (45--60\,dB), and high (70--75\,dB), representing quiet indoor, office, and crowded environments, respectively. Noise levels were measured using a sound level meter, and the same speech commands were tested under each condition \cite{kim2022robustspeech}.

\subsection{Evaluation Metrics}

LUMO was evaluated using Word Error Rate (WER), response latency, memory usage, power consumption, and thermal stability. Speech recognition performance was measured as

\begin{equation}
WER=\frac{S+D+I}{N},
\end{equation}

where $S$, $D$, and $I$ denote substitution, deletion, and insertion errors, respectively, and $N$ is the total number of reference words. Lower WER indicates better recognition performance.

End-to-end response latency was measured from the end of user speech to the start of synthesized audio:

\begin{equation}
T_{\mathrm{total}}=
T_{\mathrm{ASR}}+
T_{\mathrm{LLM}}+
T_{\mathrm{TTS}},
\end{equation}

where the terms represent ASR, language-model inference, and speech-synthesis latency, respectively. Memory usage, power consumption, and CPU temperature were monitored during continuous inference to evaluate resource efficiency and thermal stability \cite{han2024edgeenergy}.

\subsection{Privacy Preservation and Cloud Independence}

Cloud independence was evaluated using network ingress/egress volume ($V_{net}$) and external data exposure time ($T_{exp}$):

\begin{equation}
V_{net}=B_{in}+B_{out},
\end{equation}

where $B_{in}$ and $B_{out}$ represent transmitted and received network data, respectively.

\begin{equation}
T_{exp}=t_{return}-t_{transit},
\end{equation}

where $T_{exp}$ represents the duration for which user data remains outside the local device. Since LUMO performs ASR, LLM inference, and TTS entirely on-device without external APIs or cloud storage, both metrics remain zero during offline operation ($V_{net}=0$ and $T_{exp}=0$).

\subsection{Baseline Systems for Comparison}

LUMO was conceptually compared with Rhasspy and Mycroft, two open-source voice-assistant frameworks. These systems primarily use predefined intents, command grammars, or rule-based processing, whereas LUMO combines offline speech processing with a quantized local LLM for more flexible and context-aware interaction \cite{wu2023speechllm}.

\section{Results and Discussion}

This section evaluates LUMO in terms of speech recognition, multilingual capability, lightweight LLM performance, response latency, power consumption, thermal stability, and offline operation on Raspberry Pi 5.

\subsection{Offline Speech Recognition Performance}

Speech recognition was evaluated using Word Error Rate under different environmental noise levels.

\begin{table}[h]
\centering
\caption{Speech Recognition Performance Under Noise Levels}
\label{tab:wer_results}
\footnotesize
\setlength{\tabcolsep}{6pt}
\renewcommand{\arraystretch}{1.0}
\begin{tabular}{c c c}
\hline
\textbf{Noise Level} & \textbf{Environment} & \textbf{WER (\%)} \\
\hline
30 dB & Quiet room & 4.1 \\
50 dB & Office noise & 6.8 \\
70 dB & Public environment & 14.2 \\
\hline
\end{tabular}
\end{table}

Table~\ref{tab:wer_results} shows that WER increased with environmental noise.

\begin{figure}[h]
\centering
\includegraphics[width=0.8\columnwidth]{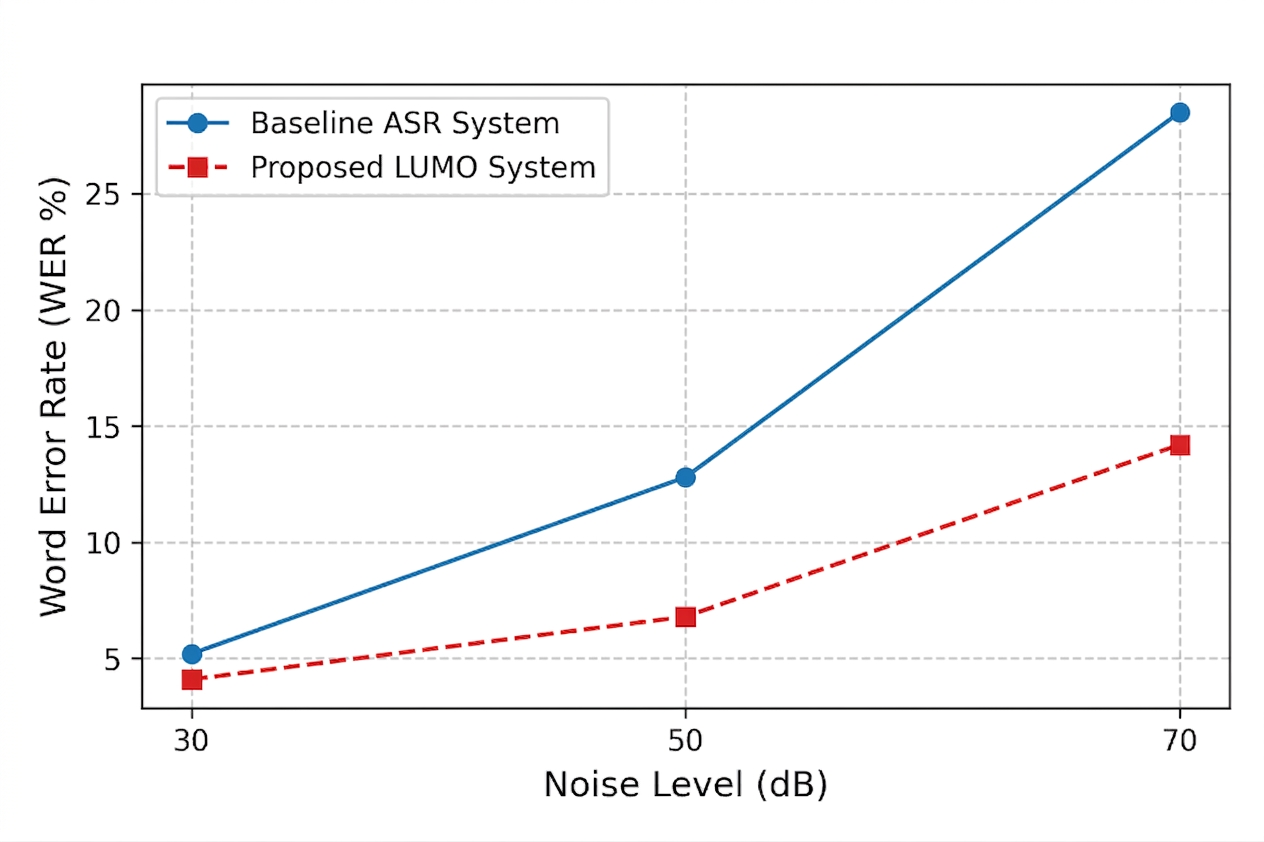}
\caption{Comparison of Word Error Rates under different noise conditions.}
\label{fig:wer_graph}
\end{figure}

As shown in Fig.~\ref{fig:wer_graph}, WER was 4.1\% at 30\,dB, increased to 6.8\% at 50\,dB, and reached 14.2\% at 70\,dB.The results show strong recognition in quiet and moderate noise conditions and deterioration in high noise. Similar behavior has been observed for Kaldi-based embedded ASR systems \cite{povey2011kaldi}.

\subsection{Multilingual Performance Analysis}

LUMO was tested on spoken commands in English and Bangla.WER was better in English but Bangla had a slightly higher error rate as expected since the training resources are limited for low resource languages \cite{pratap2020massively}.

\begin{table}[h]
\centering
\caption{Multilingual Speech Recognition Results}
\label{tab:multilingual_results}
\footnotesize
\setlength{\tabcolsep}{7pt}
\renewcommand{\arraystretch}{1.0}
\begin{tabular}{c c c}
\hline
\textbf{Language} & \textbf{Avg. Utterance Length} & \textbf{WER (\%)} \\
\hline
English & 3--5 words & 6.8 \\
Bangla & 3--5 words & 9.9 \\
\hline
\end{tabular}
\end{table}

The results show that LUMO supports English and Bangla speech recognition and works totally offline.

\subsection{Lightweight LLM Benchmark and Model Selection}

Several lightweight language models were evaluated for edge deployment based on inference speed, memory consumption, and reasoning capability. TinyLLaMA was selected because it provided the most suitable balance between inference speed and memory requirements \cite{zhang2024tinyllama}.

\begin{table}[h]
\centering
\caption{Lightweight LLM Benchmark on Raspberry Pi 5}
\label{tab:llm_benchmark}
\footnotesize
\setlength{\tabcolsep}{3.5pt}
\renewcommand{\arraystretch}{1.0}
\begin{tabular}{c c c c}
\hline
\textbf{Model} & \textbf{Parameters} & \textbf{Tokens/s} & \textbf{Memory} \\
\hline
TinyLLaMA & 1.1B & 12.3 & 640 MB \\
Gemma-2B & 2B & 4.5 & 1.4 GB \\
Phi-3 Mini & 3.8B & 2.8 & 2.2 GB \\
\hline
\end{tabular}
\end{table}

As shown in Table~\ref{tab:llm_benchmark}, TinyLLaMA achieved the highest inference speed and lowest memory consumption among the evaluated models.

\subsection{Quantization Impact on Memory and Speed}

Quantization reduces model memory requirements and improves inference efficiency on Raspberry Pi. Lower-precision formats substantially reduced model size while maintaining practical execution performance \cite{dettmers2023qlora}.

\begin{table}[h]
\centering
\caption{Impact of Quantization on Model Size}
\label{tab:quantization}
\footnotesize
\setlength{\tabcolsep}{7pt}
\renewcommand{\arraystretch}{1.0}
\begin{tabular}{c c c}
\hline
\textbf{Precision} & \textbf{Model Size} & \textbf{Reduction} \\
\hline
FP16 & 2.2 GB & -- \\
8-bit & 1.1 GB & 50\% \\
4-bit GGUF & 640 MB & 72\% \\
\hline
\end{tabular}
\end{table}
Table~\ref{tab:quantization} shows that 4-bit GGUF was the largest reducer, reducing the model size by 72\% in comparison with FP16.

\subsection{End-to-End Latency Analysis}

End-to-end latency was measured from when the user stopped talking to when synthesized audio playback started.

\begin{table}[h]
\centering
\caption{Latency Breakdown of LUMO Processing Pipeline}
\label{tab:latency_breakdown}
\footnotesize
\setlength{\tabcolsep}{9pt}
\renewcommand{\arraystretch}{1.0}
\begin{tabular}{c c}
\hline
\textbf{Processing Stage} & \textbf{Latency (s)} \\
\hline
VAD + STT & 0.7 \\
LLM Inference & 1.8 \\
TTS Synthesis & 0.7 \\
\textbf{Total} & \textbf{3.2} \\
\hline
\end{tabular}
\end{table}

\begin{figure}[h]
\centering
\includegraphics[width=0.8\columnwidth]{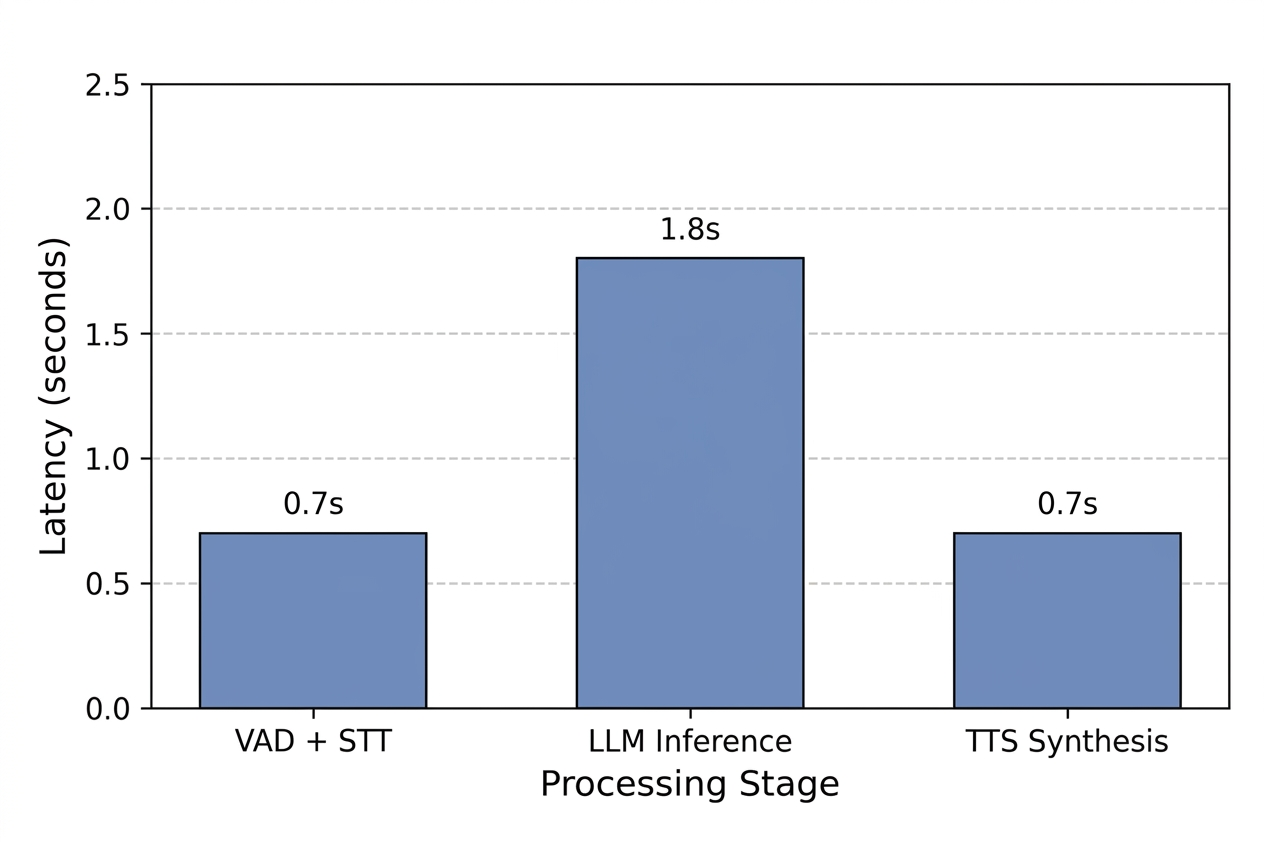}
\caption{Latency contributions of different stages in the LUMO pipeline.}
\label{fig:latency}
\end{figure}

The latency of VAD+STT and TTS was 0.7\,s each, and that of LLM inference was 1.8\,s, which accounted for 56.3\% of the total response time of 3.2\,s, as shown in Fig.~\ref{fig:latency}. Therefore, the dominant computational bottleneck is LLM inference.Similar patterns of latency have been observed in mobile and embedded deep-learning systems \cite{lane2017squeezing}.

\subsection{Power Consumption Analysis}

The power consumption was monitored continuously during system operation. Results show that LUMO works at a power range that is suitable for edge computing \cite{han2024edgeenergy}.

\begin{table}[h]
\centering
\caption{Power Consumption During System Operation}
\label{tab:power_consumption}
\footnotesize
\setlength{\tabcolsep}{10pt}
\renewcommand{\arraystretch}{1.0}
\begin{tabular}{c c}
\hline
\textbf{System State} & \textbf{Power (W)} \\
\hline
Idle & 4.2 \\
Speech Recognition & 6.8 \\
LLM Inference & 9.0 \\
\hline
\end{tabular}
\end{table}

LUMO consumed 4.2\,W at idle, 6.8\,W during speech recognition and 9.0\,W during LLM inference.

\begin{figure}[h]
\centering
\includegraphics[width=0.8\columnwidth]{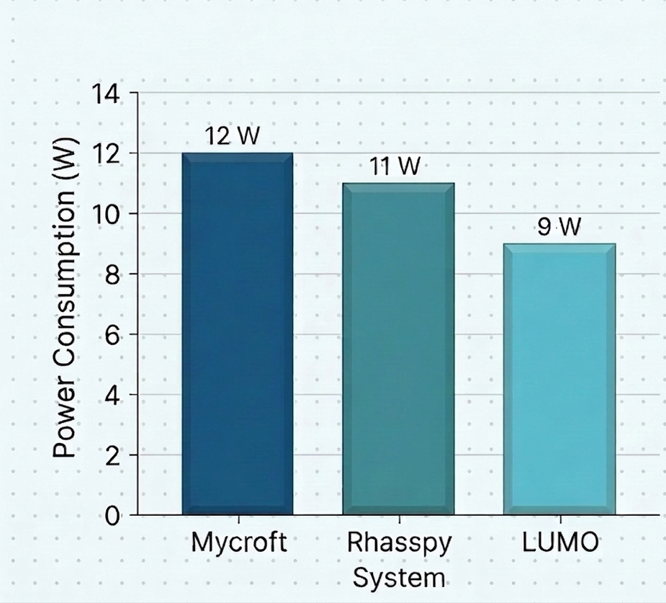}
\caption{LUMO vs. Other Offline Assistants: Power Consumption Comparison}
\label{fig:power}
\end{figure}
LUMO consumed 9\,W (vs. about 11\,W for Rhasspy and 12\,W for Mycroft, see Fig.~\ref{fig:power}).This equates to reductions of 18.2\% and 25.0\%, respectively.

\subsection{Thermal Stability Analysis}

We also monitored the CPU temperature during LLM inference continuously to assess the thermal stability.
\begin{figure}[h]
\centering
\includegraphics[width=0.8\columnwidth]{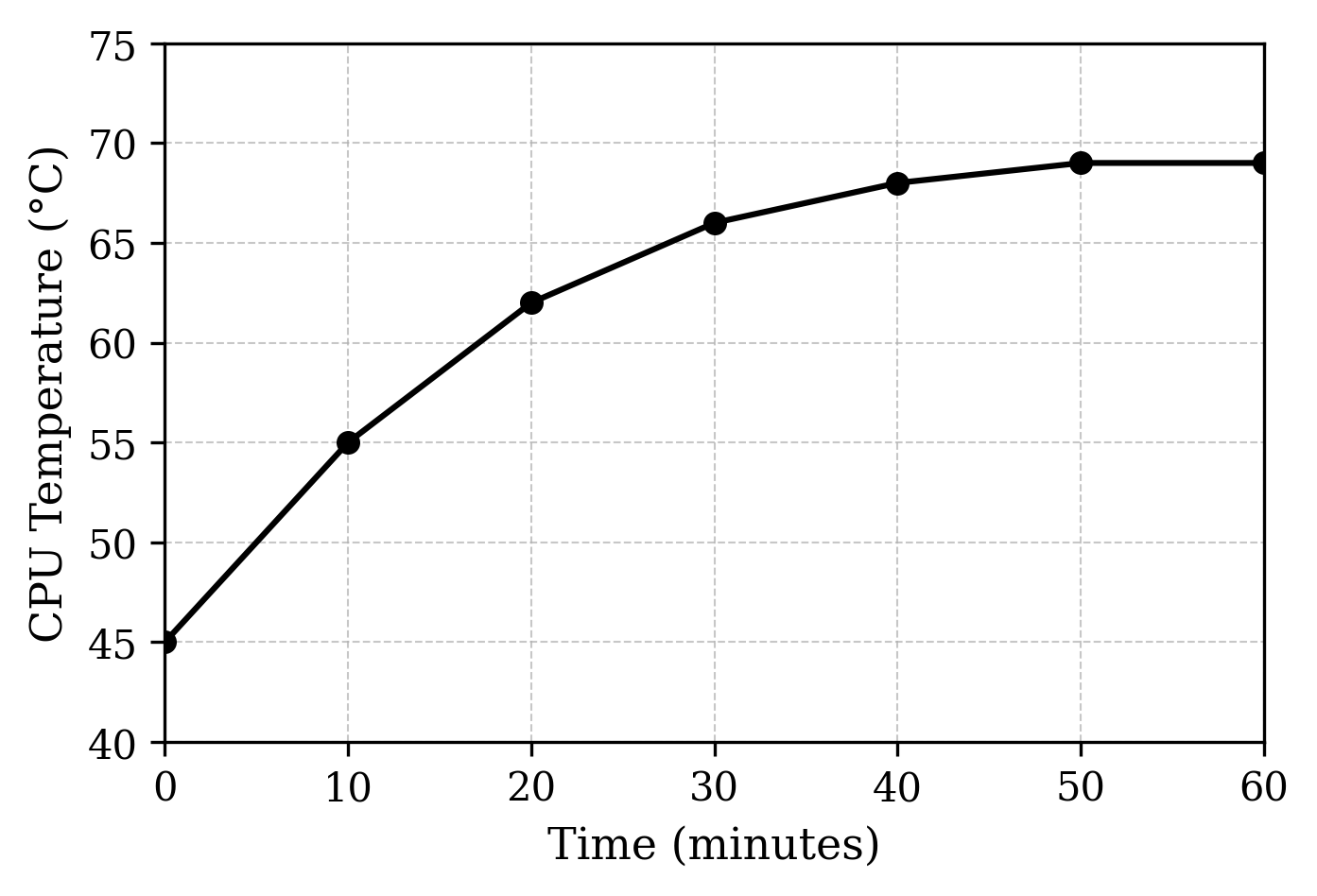}
\caption{CPU temperature change for continual LLM inference.}
\label{fig:temperature}
\end{figure}

As shown in Fig.~\ref{fig:temperature}, the temperature of the CPU increased gradually during the LLM inference, but it was below $70^\circ$C, which indicates the stable operation of the CPU without any sudden thermal degradation. Temperature control is required for reliable embedded artificial intelligence systems \cite{shi2016edge}.

\subsection{Comparison with Rhasspy and Mycroft}

LUMO was compared with Rhasspy and Mycroft to assess privacy, connectivity, latency, power consumption, and local language-model support.

\begin{table}[h]
\centering
\caption{Comparison Based on Data Privacy and Cloud Connectivity}
\label{tab:privacy_comparison}
\footnotesize
\setlength{\tabcolsep}{2.5pt}
\renewcommand{\arraystretch}{1.0}
\begin{tabular}{c c c c c}
\hline
\textbf{System} & \textbf{Processing} & $\mathbf{V}_{net}$ & \textbf{Medium} & \textbf{Risk} \\
\hline
Mycroft & Local/Cloud & Variable & Local+Cloud & Partial \\
Rhasspy & Fully Local & Minimal & Local Only & Low \\
\rowcolor{gray!20}
LUMO & On-Device & 0.0 KB & None & Zero \\
\hline
\end{tabular}
\end{table}

Table~\ref{tab:privacy_comparison} shows that LUMO operates entirely on-device with zero outbound traffic during offline execution, preventing conversational data from being transmitted to external services.

\begin{table}[h]
\centering
\caption{Comparison with Existing Offline Assistants}
\label{tab:assistant_comparison}
\footnotesize
\setlength{\tabcolsep}{3.5pt}
\renewcommand{\arraystretch}{1.0}
\begin{tabular}{c c c c c}
\hline
\textbf{System} & \textbf{Offline} & \textbf{LLM} & \textbf{Latency} & \textbf{Power} \\
\hline
Mycroft & Partial & No & $\sim$5 s & 12 W \\
Rhasspy & Yes & No & $\sim$3.5 s & 11 W \\
\rowcolor{gray!20}
LUMO & Yes & Yes & 3.2 s & 9 W \\
\hline
\end{tabular}
\end{table}

Table~\ref{tab:assistant_comparison} shows that LUMO achieved 3.2\,s latency, improving by 36.0\% over Mycroft and 8.6\% over Rhasspy. Its 9\,W power consumption was also lower than Rhasspy (11\,W) and Mycroft (12\,W). Unlike the compared systems, LUMO supports local generative LLM-based reasoning for more flexible conversational interaction.

LUMO's performance is supported by 4-bit GGUF quantization, TinyLLaMA, multithreaded processing, and fully offline execution, resulting in 3.2\,s response latency and 9\,W peak power consumption. WebRTC VAD, VOSK ASR, and the command-style dataset support low speech-recognition error rates. Current limitations include performance in highly noisy environments, limited advanced reasoning, voice-only interaction, and the absence of IoT context and long-term memory. Future work will investigate broader multilingual evaluation, newer edge language models, and sensor-based context-aware interaction.
\section{Conclusion}

This paper presented LUMO (Lightweight Unified Multilingual Orchestrator), a privacy-preserving offline voice assistant for resource-constrained edge devices that integrates offline ASR, a quantized lightweight language model, and text-to-speech synthesis on a Raspberry Pi. By eliminating cloud dependence, LUMO supports private and reliable operation in connectivity-limited environments. Experimental results demonstrated reliable speech recognition under varying noise conditions, practical response latency, and stable power and thermal performance, supporting its feasibility for edge-AI applications.
There are still some limitations such as limited reasoning capability of the compact language model, comparatively higher Bangla WER and limited evaluation on the longer multi-turn conversations.In conclusion, we show that privacy-preserving conversational assistants can be successfully deployed on low-power edge devices through model optimization and efficient pipeline integration.Future work will focus on improvements in Bangla ASR, evaluation of newer light-weight reasoning models, and evaluation of conversational scalability under edge hardware constraints for further extension of LUMO for edge-based conversational AI.

\bibliographystyle{IEEEtran}
\bibliography{biblio}

\end{document}